\documentclass{article}

\usepackage{microtype}
\usepackage{graphicx}
\usepackage{subfigure}
\usepackage{booktabs} 
\usepackage{amsmath}
\usepackage{amssymb}
\usepackage{subcaption}
\usepackage{multirow}
\usepackage{multicol}
\usepackage{threeparttable}

\usepackage{hyperref}

\usepackage[accepted]{mlsys2025}
\usepackage{xspace}
\newcommand{\sys}{\textsc{Prism}\xspace}

\makeatletter
\def\isaccepted{1}
\makeatother

\usepackage{hyperref}
\mlsystitlerunning{PRISM: Parametrically Refactor Inference for Speculative decoding draft Models}

\begin{document}

\twocolumn[
\mlsystitle{\includegraphics[height=2em]{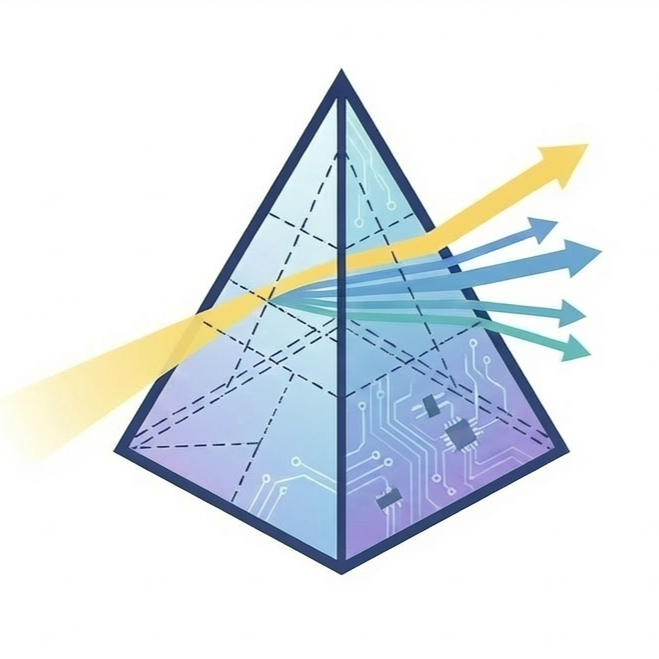} PRISM: \underline{P}arametrically  \underline{R}efactoring  \underline{I}nference for  \underline{S}peculative Sampling Draft  \underline{M}odels}

\mlsyssetsymbol{equal}{*}

\begin{mlsysauthorlist}
\mlsysauthor{Xuliang Wang}{equal,uw,cra}
\mlsysauthor{Yuetao Chen}{equal,cuhk}
\mlsysauthor{Maochan Zhen}{cra}
\mlsysauthor{Fang Liu}{cra}
\mlsysauthor{Xinzhou Zheng}{utsc}
\mlsysauthor{Xingwu Liu}{dlut,cra}
\mlsysauthor{Hong Xu}{cuhk}
\mlsysauthor{Ming Li}{uw,cra}
\end{mlsysauthorlist}

\mlsysaffiliation{uw}{David R. Cheriton School of Computer Science, University of Waterloo, Waterloo, Ontario, Canada}
\mlsysaffiliation{cuhk}{The Chinese University of Hong Kong, Hong Kong SAR, China}
\mlsysaffiliation{cra}{Central China Institute of Artificial Intelligence, Zhengzhou, Henan, China}
\mlsysaffiliation{dlut}{Dalian University of Technology, Dalian, Liaoning, China}
\mlsysaffiliation{utsc}{University of Science and Technology of China, Hefei, Anhui, China}

\mlsyscorrespondingauthor{Ming Li}{mli@uwaterloo.ca}

\mlsyskeywords{Machine Learning, MLSys}

\vskip 0.3in

\begin{abstract}
Large Language Models (LLMs), constrained by their auto-regressive nature, suffer from slow decoding. Speculative decoding methods have emerged as a promising solution to accelerate LLM decoding, attracting attention from both systems and AI research communities. Recently, the pursuit of better draft quality has driven a trend toward parametrically larger draft models, which inevitably introduces substantial computational overhead. While existing work attempts to balance the trade-off between prediction accuracy and compute latency, we address this fundamental dilemma through architectural innovation.

We propose \sys, which disaggregates the computation of each predictive step across different parameter sets, refactoring the computational pathways of draft models to successfully decouple model capacity from inference cost. Through extensive experiments, we demonstrate that \sys outperforms all existing draft architectures, achieving exceptional acceptance lengths while maintaining minimal draft latency for superior end-to-end speedup. We also re-examine scaling laws with \sys, revealing that \sys scales more effectively with expanding data volumes than other draft architectures. Through rigorous and fair comparison, we show that \sys boosts the decoding throughput of an already highly optimized inference engine by more than 2.6×.
\end{abstract}
]

\printAffiliationsAndNotice{\mlsysEqualContribution} 

\section{Introduction}
\label{introduction}

Large Language Models (LLMs) have become a foundational technology. Models such as OpenAI's ChatGPT~\cite{OpenAI2022ChatGPT}, Meta's LLaMA~\cite{Touvron2023LLaMA}, and Google's Gemini~\cite{Gemini2023_leads} have demonstrated remarkable capabilities across a diverse range of tasks, proving potentials for novel applications that advance human-computer interaction, optimize industrial workflows, and create new paradigms for scientific endeavors. However, realizing these potentials in practical applications is often constrained by the critical problem of generation efficiency.

The efficiency problem is inherent to the auto-regressive nature of LLM inference. Auto-regressive generations usually adopt a sequential scheme, performing a full forward pass through its parameters to produce each token, which results in substantial computational cost and low throughput.

To mitigate this problem, the research community has proposed a range of acceleration techniques. While optimizations such as efficient KV cache management~\cite{Kwon2023Efficient, 305212}, continuous batching~\cite{Yu2022Orca}, parallelism optimization~\cite{Li2023AlpaServe, 10.1109/SC41406.2024.00046}, and prefill-decode disaggregation schemes~\cite{Zhong2024TetriInfer, 10.5555/3691938.3691949, 10.1109/ISCA59077.2024.00019} have yielded significant speedups, they do not alter the fundamental token-by-token generation paradigm. As a complementary approach, speculative decoding~\cite{leviathan2023fast} effectively reduces the number of LLM forward passes.

Speculative decoding, also known as Draft-and-Verify, is inspired by speculative execution, a well-established optimization technique in modern CPU design. It primarily consists of two stages: in the first stage, a smaller, faster draft model proposes a sequence of candidate tokens; in the subsequent stage, the drafted sequence is verified by the larger, slower target model. Intuitively, when multiple drafted tokens are accepted, speculative decoding enables the target model to generate multiple tokens in a single forward pass, whereas naive auto-regressive decoding generates only one token per pass. Consequently, the success of speculative decoding strongly relies on draft quality, i.e. , how many tokens proposed by the draft model are expected to be accepted by the target model in one draft.

To improve draft quality, many drafter models, following EAGLE~\cite{li2024eagle}, utilize the target model's intermediate hidden states from the previous verification step as additional input. This enables more accurate drafts than drafter models that rely on tokens alone. Many of these draft models employ only one transformer block to minimize the computational overhead of drafting.

\begin{table}[t]
\caption{A comparison of draft model relative sizes used in related works, measured in drafter parameter counts divided by target model sizes. Note that the statistics include all parameters activated in the forward pass and LLMs mentioned is the chat/Instruct Version. This table shows a growing trend in adopting relatively larger drafter models.}
\label{tab:large drafter}
\vskip 0.1in
\centering
\scriptsize
\begin{threeparttable}
\begin{tabular}{lllc}
\toprule
Paper & Base LLM & Drafter  & Relative Size \\
\midrule
\citeauthor{leviathan2023fast} & T5-XXL-11B & T5-SMALL (77M) & 0.68\%\\
\citeauthor{10.1145/3620666.3651335} & LLaMA-2-7B & LLaMA-68M & 0.95\% \\
 & LLaMA-2-65B & LLaMA-68M & 0.10\% \\
\midrule
\citeauthor{li2024eagle} & LLaMA2-7B & EAGLE-2 (0.48B) & 6.86\% \\
 & LLaMA2-70B & EAGLE-2 (1.23B) & 1.76\% \\
\citeauthor{li2025eagle3scalinginferenceacceleration} & LLaMA3.1-8B & EAGLE-3 (1.03B) & 12.88\% \\
 & LLaMA3.3-70B & EAGLE-3 (2.19B) & 3.13\% \\
\citeauthor{tang2025efficientspeculativedecodingllama}\tnote{*} & LLaMA-3.3-8B & Scaled EAGLE-2 (1.59B) & 19.88\% \\ %
& Llama-3.3-70B & Scaled EAGLE-2 (4.35B) & 6.21\% \\
\citeauthor{yan2025scalinglawsspeculativedecoding}\tnote{*} & LLaMA-2-7B & Scaled EAGLE-2 (1.26B) & 18.00\% \\ 
 & LLaMA-2-13B & Scaled EAGLE-2 (1.78B) & 13.69\% \\
\bottomrule
\end{tabular}
\begin{tablenotes}
    \item[*] \scriptsize We calculate drafter sizes for these rows as the drafter models are not publicly available.
\end{tablenotes}
\end{threeparttable} 
\vskip -0.2in
\end{table}

However, the single-transformer configuration limits draft model capacity and ultimately caps acceptance rates. Recognizing this limitation, as shown in Table \ref{tab:large drafter}, emerging works have begun to explore the viability of larger draft models~\cite{li2025eagle3scalinginferenceacceleration, yan2025scalinglawsspeculativedecoding, tang2025efficientspeculativedecodingllama}. While maintaining a similar architecture, these works stacks more transformer layers. Although larger drafters do incur more computational overhead, the argument is that this cost is outweighed by the gains from improved acceptance rates.

In this paper, we investigate the fundamental trade-off between draft quality and computational overhead in speculative decoding from a different perspective. Our key insight is that while larger models are necessary for enhancing drafter predictive performance, it is possible to keep draft overhead low simultaneously. We demonstrate this through \sys, which recognizes the inherent differences between draft steps and leverages conditional computing. Specifically, we refactor the inference computational path by distributing computation across different draft steps to distinct parameter sets, analogous to how a prism disperses white light into its spectrum. Consequently, the total parameter count of the drafter expands while the number of activated parameters per draft step remains constant. Fewer activated parameters accelerate both inference and training. Critically, the model's representational capacity still expands: through extensive experiments, we show that \sys's architecture not only preserves favorable scaling properties but actually scales more effectively than naive parameter scaling approaches such as vertically stacking transformer layers.

In summary, this paper addresses the core trade-off in speculative decoding between drafter predictive quality and efficiency by introducing \sys, a novel drafter architecture. Our contributions include:

\begin{itemize}
    \item \textbf{A Novel Architecture that Decouples Draft Model Capacity from Draft Computational Cost.} We introduce \sys, the first architecture to apply a conditional computing paradigm specifically across auto-regressive generation steps to expand model capacity. Unlike conventional draft models where capacity and computational overhead are entangled, \sys successfully decouples them by parametrically disaggregating the computation of different draft steps. This novel architecture propose new perspective for drafter effectiveness-efficiency trade-off, showing new paths for drafter design.

    \item \textbf{Re-examining the Scaling Law for Draft Models.} Our work provides the first empirical evidence that a draft model's predictive power can scale effectively without increasing the activated parameter count. This finding establishes a new, more efficient scaling paradigm for draft models.

    \item \textbf{Detailed Evalution with Inference Engine.} We bridge the gap between speculative decoding optimizations from the AI and the system academia community by implementing and validating \sys within SGLang, a state-of-the-art inference engine. In contrast to many prior drafters evaluated solely in PyTorch-based settings, our experiments provide robust evidences of \sys's efficacy with a system-level highly optimized inference engine.
\end{itemize}

\section{background}
This section provides a brief overview of LLM inference, speculative decoding and state-of-the-art drafter architectures.

\textbf{LLM Inference.} LLM inference is fundamentally auto-regressive, probabilistically generating a sequence of tokens $T = (t_1, t_2, ..., t_n)$ where each token $t_i$ is sampled from a conditional probability distribution parameterized by the model weights $\theta$ and the preceding tokens: $p(t_i | t_1, ..., t_{i-1}; \theta)$. This sequential dependency makes an iterative generation process necessary.

By computaional characteristics, LLM inference are bifurcated into two distinct phases, prefill and decoding.

The prefill phase refers to the initial phase involves the processing of the input prompt. The model performs a highly parallel forward pass to compute the attention mechanism's key-value (KV) pairs for all prompt tokens simultaneously. This phase is computation-bound, contributed by large matrix-matrix multiplications across the entire input sequence.

Following the prefill, the model enters the iterative decoding phase. Different from the prefill phase, the decoding phase is memory-bandwidth-bound as it necessitates loading the entire set of model parameters from off-chip high-bandwidth memory (HBM) to the on-chip processing units for each step. This memory-bandwidth bottleneck is further exacerbated in scenarios involving long contexts where with each generated token, the model must also read the attention KV caches for all preceding tokens.

\textbf{Speculative Decoding.}
The core principle of Speculative Decoding involves using a parametrically smaller, faster draft model $M_{draft}$ to generate a sequence of candidate tokens based on the current context, which are then parallelly validated in a single forward pass by the parametrically larger, slower target model $M_{target}$. 

Formally, let $p_{target}(t|c)$ be the probability distribution over the next token $t$ given a context $c$, as defined by $M_{target}$. Similarly, let $p_{draft}(t|c)$ be the distribution from $M_{draft}$. During drafting, At a given step with context $c$, $M_{draft}$ is invoked to generate a candidate sequence of $\gamma$ tokens, $\tilde{T} = (\tilde{t}_1, \tilde{t}_2, ..., \tilde{t}_\gamma)$. Then for verification, $M_{target}$ performs a single forward pass on the concatenated input $(c, \tilde{T})$ to efficiently compute the true probability distributions for each position: $p_{target}(\cdot|c)$, $p_{target}(\cdot|c, \tilde{t}_1)$, ..., $p_{target}(\cdot|c, \tilde{t}_1, ..., \tilde{t}_{\gamma-1})$. The draft tokens are validated sequentially using a modified rejection sampling scheme: for each draft token $\tilde{t}_i$, where the initial value of $i$ is 1, it is accepted if a randomly drawn number $r \in \text{Uniform}(0, 1)$ which satisfies
$$
    r < \min (1,\frac{p_{target}(\tilde{t}_i|c,\tilde{t}_{<i})}{p_{draft}(\tilde{t}_i|c,\tilde{t}_{<i})})
$$
The process iterates over $i$, until all tokens are accepted or a token $\tilde{t}_k$ is rejected. All accepted tokens $\hat{T}$ are kept. Additionally, an extra token could be sampled from a corrected distribution $p'_{target} \propto \max(0, p_{target}(\cdot|c,\hat{T}) - p_{draft}(\cdot|c,\hat{T}))$. Following this, the model's context is updated with the newly generated tokens, the KV cache entries corresponding to any rejected candidates are discarded, and a new drafting cycle commences.

A fundamental and critical property of speculative decoding is that it is lossless. This has been formally proven and guaranteed that the final output sequence produced by the speculative process is drawn from the exact same probability distribution as if it were generated token-by-token by $M_{target}$ alone~\cite{10.1145/3620666.3651335}.

From the system view, the efficacy of speculative decoding lies in its ability to amortize the high cost of the memory-bandwidth-bound decoding steps. The conventional one-token-per-pass auto-regressive scheme is heavily memory-bandwith-bound, leading to poor GPU utilization as the computing units stall waiting for data. Speculative decoding transforms the workload by using a single forward pass to verify multiple tokens in parallel. When the draft model's predictions align well with the target model's, leading to a high acceptance rate, the effective cost per token is drastically reduced. This mechanism fundamentally alters the operational characteristics of the decoding phase, increasing its compute-to-memory ratio.

\textbf{SoTA Draft Models.}
Initial approaches to speculative decoding primarily utilized lightweight versions of the target LLM as draft models. A significant difference was introduced by EAGLE~\cite{li2024eagle}, who first leverages not only the preceding context tokens but also the hidden states generated by the target model during the most recent verification step. By conditioning on this more informative input, EAGLE's draft models can more accurately predict the target model's distribution, leading to a substantially higher acceptance rate for candidate tokens. The efficacy of exploiting informative intermediate representations from the target model has led to its adoption in subsequent works~\cite{li2024eagle2fasterinferencelanguage, li2025eagle3scalinginferenceacceleration, huang2025jakiroboostingspeculativedecoding, zhang2025learning}. 

Most SoTA draft models also utilize a tree-based Draft-and-Verify approach instead of the naive sequence-based approach. Initially proposed by Specinfer~\cite{10.1145/3620666.3651335}, the key observation is that even when the most likely drafted token is rejected, an alternative, less probable candidate from the same generation probability distribution might still be correct. By allowing multiple tokens to be validated for the same generation step, the draft tokens form different candidate paths and can be organized into a tree structure. During the verification phase, a special tree-shaped attention mask is applied to enforce the causal relationships among the draft tokens. Tree-based draft and verify significantly enhances acceptance rate.

\textbf{Scaling of drafter models.}
The expectation of draft models has previously been governed by the principle of minimizing computational overhead, resulting in parametrically small drafters. This paradigm, however, inherently limits the model's capacity and its ability to benefit from large-scale training data. Progressively recognizing the limitation, recent research has shifted towards balancing the trade-off between draft model complexity and its computation latency. The goal is to identify a \emph{sweet spot} where a more powerful, albeit more costly, draft model that yields a high enough acceptance rate to deliver high end-to-end acceleration. 

EAGLE-3~\cite{li2025eagle3scalinginferenceacceleration} pioneered the study of intentionally scaling draft models, demonstrating that increasing trainable parameters and making other architectural refinements enabled considerable performance gains when trained with more data; Scylla~\cite{yan2025scalinglawsspeculativedecoding} scaled the EAGLE-2 architecture with up to five trainable transformer layers, pretraining and tuning it on a considerably larger dataset, resulting in state-of-the-art acceptance lengths. In parallel, Meta~\cite{tang2025efficientspeculativedecodingllama} has explored different axes of scaling, investigating methods for parameter expansion such as increasing model depth and substituting standard Feed-Forward Networks (FFN) with Mixture-of-Experts (MoE) layers. These efforts collectively signal a transfer of expectations for draft models, moving from the minimal overhead to balanced performance optimized for maximum system throughput.

\section{methodology}
\subsection{Motivation}
As the practicability of scaling draft models to achieve high acceptance rates emerges, it is compelling to train draft models with considerably more parameters. To expand model capacity without introducing extra overhead, we drew inspiration from the MoE models, which leverage specialization by routing computations to different sub-networks. We observe a similar opportunity for specialization exists within speculative decoding: the predictive task in speculative decoding is non-uniform in its difficulty. There is a clear empirical trend where drafting becomes progressively harder at later steps in the sequence. This is consistently demonstrated by the sharp decline in acceptance rates for later tokens~(see Figure \ref{fig:ar})
. This insight directly motivates our proposed architecture, \sys. Rather than relying on a single, one-size-fits-all model, \sys employs a principle of specialization by applying different set of parameters to different draft steps. This design brings two benefits intuitively:

\begin{figure}[tb]
    \centering
    \includegraphics[width=0.9\linewidth]{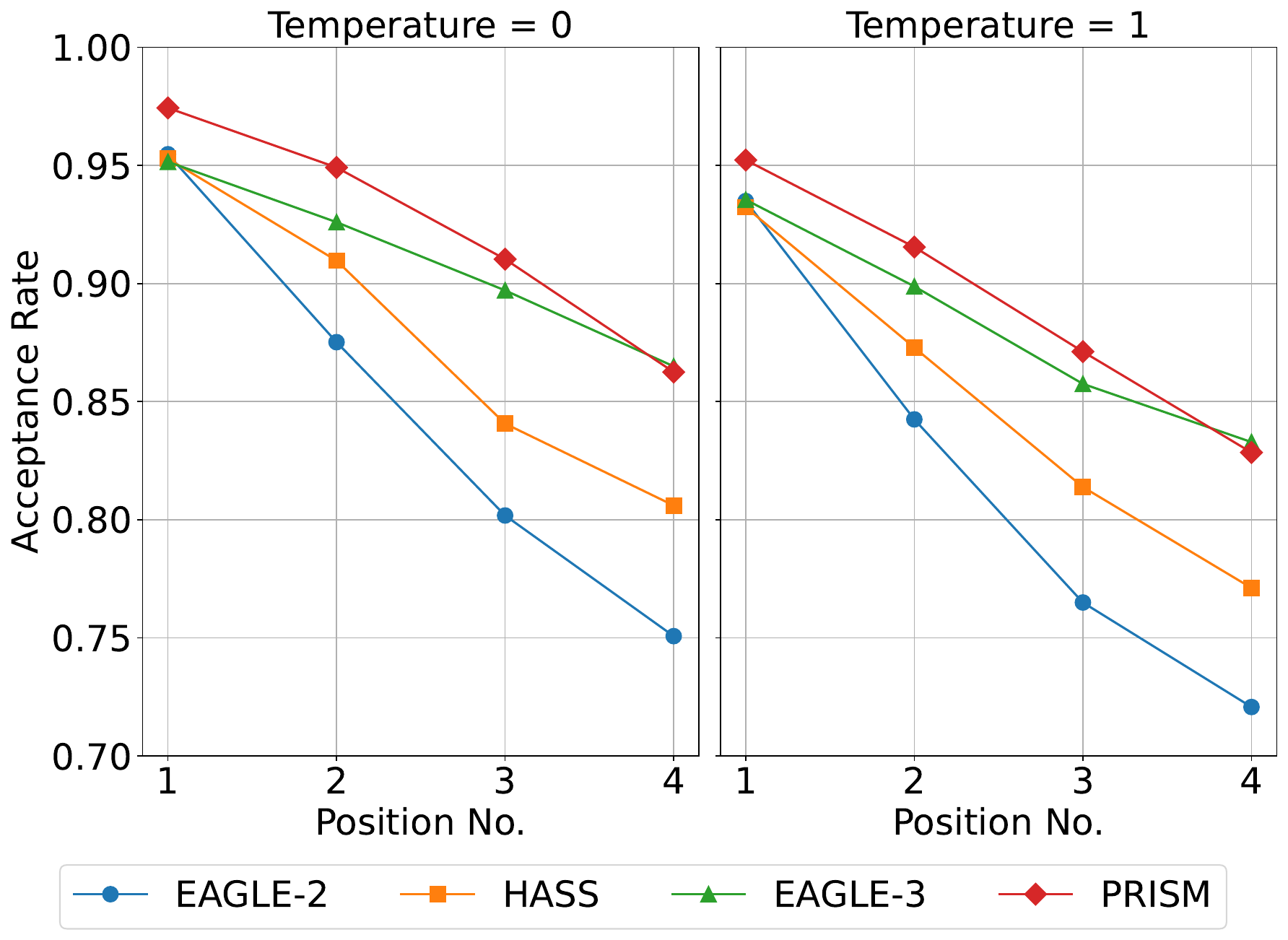}
    \caption{Emperical results showing the average acceptance rate step-wise on LLaMA-3-8B.}
    \label{fig:ar}
    \vskip -0.2in
\end{figure}

\begin{figure}
    \centering
    \includegraphics[width=0.9\linewidth]{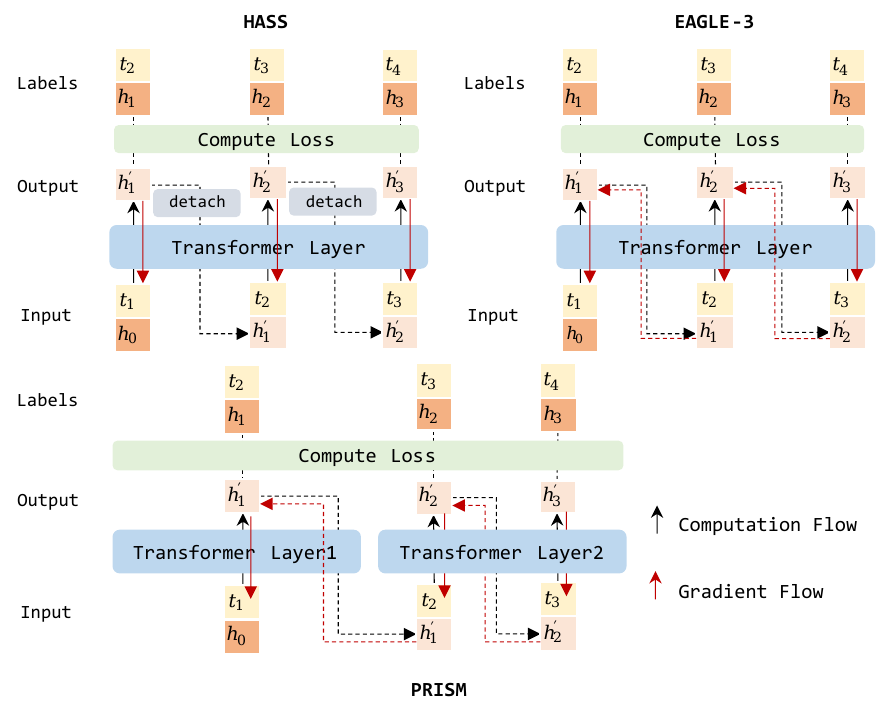}
    \caption{An illustration of computation and gradient flow of HASS, EAGLE-3 and \sys.}
    \label{fig:pr}
    \vskip -0.1in
\end{figure}

\begin{itemize}
    \item \textbf{Adaptive Representation Complexity.} \sys effectively creates a hierarchical model where later parameters process the output of previous ones, forming a cascaded computational structure. This design enables adaptive allocation of computational resources across draft steps. As shown in Figure \ref{fig:pr}, compared with previous architectures where all draft steps utilize the same set of parameters across draft steps, \sys progressively increases the effective depth with draft steps. This architectural choice naturally allocates more cumulative computation to harder prediction tasks at the latter part of the sequence, where uncertainty typically increases and accurate token prediction becomes more challenging.
    
    \item \textbf{Decoupling Model Capacity from Inference Cost.} Critically, this architecture decouples the model's total learning capacity from its per-step inference cost. We can substantially increase the total number of trainable parameters, allowing the model to learn more complex functions, while the computational cost of generating any single draft token remains constant and minimal.
\end{itemize}

\subsection{\sys Architecture}

\begin{figure*}
    \centering
    \includegraphics[width=0.95\textwidth]{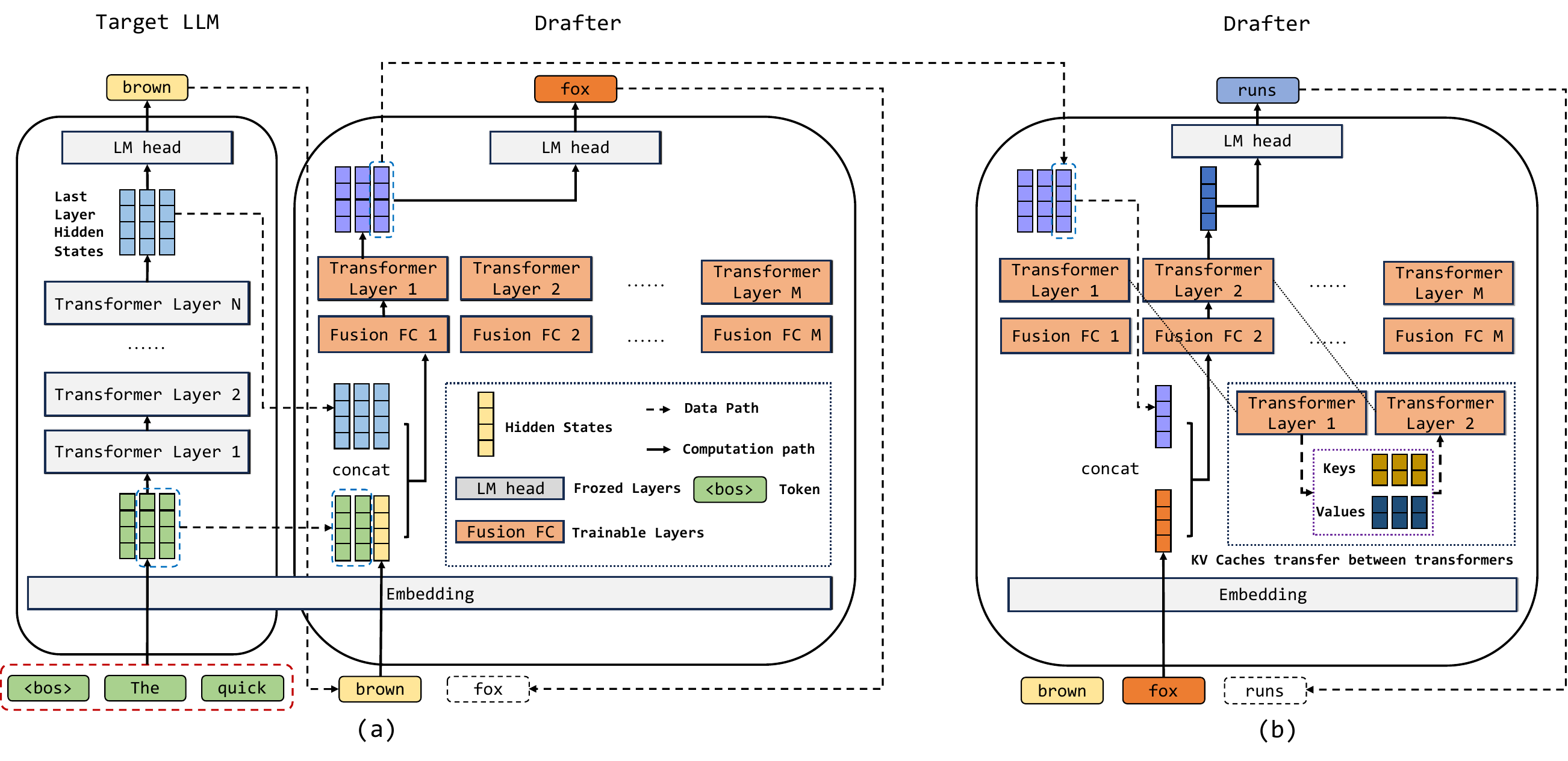}
    \vspace{-8pt}
    \caption{An illustration of the \sys architecture and its computing paths. (a) The target model prefills or verifies multiple tokens and the output hidden states of its last-layer transformer is used as an input for the draft model, fused with the input token embedding. The fused features is fed to the first transformer of the \sys drafter for prefill. (b) After fusing the embedding of the previous generated token and the previous generated hidden states, apply the fused feature to the second transformer layer for the first decoding step. Note that KV caches from transfomer 1 is transfered to transformer 2.}
    \label{fig:arc}
\end{figure*}

\sys, or Parametrically Refactor Inference for Speculative Decoding draft Models, follows the SoTA Drafters' convention to be a transformer-based drafter. As shown in Figure~\ref{fig:arc}, \sys consists of a token embedding layer, a final vocabulary projection head and a sequence of processing modules. Each processing module consists of a fully connected fusion layer and a transformer layer. 

The relationship between draft steps and processing modules in \sys is defined by a surjection. We establish the surjection by assigning each draft step to a specific processing module for its computation. This mapping allows for a many-to-one relationship: while every step has a uniquely assigned module, a single processing module can serve multiple steps.

\begin{table}[t]
\caption{Notation for \sys components and variables}
\label{tab:notation}
\vskip 0.1in
\begin{center}
\begin{scriptsize}
\begin{tabular}{ll}
\toprule
Notation & Description \\
\midrule
$S_{acc}$ & Sequence of $n$ accepted tokens from \\
& previous verification: $(x_1, ..., x_n)$ \\
$H_{base}$ & Target model hidden states for $S_{acc}$: \\
& $(h_0, h_1, ..., h_{n-1})$ \\
\midrule
$E$ & Token embedding layer (frozen) \\
${L_1, ..., L_M}$ & $M$ distinct Transformer layers \\
$F_i(\cdot, \cdot)$ & Fusion function for $L_i$ combining \\
& token embedding and hidden state \\
$f_{map}$ & Surjection mapping $K$ draft steps \\
& to $M$ modules: ${1,...,K} \to {1,...,M}$ \\
$LM_{head}$ & Vocabulary projection layer (frozen) \\
\bottomrule
\end{tabular}
\end{scriptsize}
\end{center}
\vskip -0.1in
\end{table}

\textbf{Draft with \sys.} The draft process is featured by a unique proactive parameter switching procedure between inference forward passes, as shown in Figure~\ref{fig:arc}.

Formally, define the model's components and variables as in Table \ref{tab:notation}.

The draft starts with a prefill of the draft model. For each accepted token $x_i \in S_{acc}$, a fused representation is created by  combining its embedding with the corresponding hidden state $h_{i-1}$ from the target model and feeding it to the fusion function
$$
    H_{\text{fused}} = [F_{f_{map}(1)}(E(x_i), h_{i-1})]_{i=1}^{n}
$$
This sequence of fused representations is then processed in a single forward pass through the designated transformer layer
$$
    H'_{out}, KV_{(1)}=L_{f_{map}(1)}(H_{fused}, KV_{(0)})
$$
Where $KV_{(0)}$ is the initial state of the KV cache.The output consists of a hidden state sequence $H'_{out}$ and the updated KV cache $KV_{(1)}$. Note that the KV caches will be shared across processing modules. The final hidden state from $H'_{out}$, i.e. $h'_{n}$, becomes the initial predictive state for the drafter auto-regressive decoding phase.
$$
    \hat{h}_{0}=h'_{n}
$$
The initial input token for the decoding phase $\hat{x}_1$ is also the first draft token. It should be predicted directly from $\hat{h}_0$ using the classification head
$$
    \hat{x}_1 = \arg\max(LM_{head}(\hat{h}_0))
$$

The remaining draft steps are all decoding steps repeating similar procedures:
for $k=2,3,...,K$,
$$
    H_{in}^{(k)}, KV_{(k)}=F_{f_{map}(k)}(E(\hat{x}_{k-1},\hat{h}_{k-2}), KV_{(k-1)})
$$
$$
    \hat{h}_{k-1} = L_{f_{map}(k)}(H_{in}^{(k)})
$$
$$
    \hat{x}_k=\arg\max(LM_{head}(\hat{h}_{k-1}))
$$
The whole draft process ends with output draft tokens $(\hat{h}_{1}, \hat{h}_{2}, ..., \hat{h}_{k})$, which will then sent to be verified.

Notably, the formulation above assumes a linear generation path. This is an simplification for the sake of presentation clarity. Actually,  \sys mechanism is totally compatible with tree-structured draft and stochastic sampling methods that many other drafters equip.

\textbf{Train \sys.} To mitigate the problem of exposure bias, many drafters employ context alignment techniques during training, which significantly improves acceptance length. Prominent examples include the harmonized context alignment in HASS~\cite{zhang2025learning}, the train-time test in EAGLE-3~\cite{li2025eagle3scalinginferenceacceleration}, and token alignment in Griffin~\cite{hu2025griffineffectivetokenalignment}. Context alignment aligns input hidden states, KV caches and even input tokens from the training and the inference. With \sys, we followed the HASS style of aligning the hidden states and KV caches for 3 steps.

Context alignment simulates the auto-regressive inference process in training by requiring a separate forward pass for each draft step. This significantly increases training cost: for instance, training EAGLE-3 to draft $n$ tokens requires $n$ distinct forward passes per training sample.

\sys architecture shows advantage in training efficiency when doing context alignment. While a monolithic drafter with a comparable parameter count requires propagating activations and gradients through its entire network depth for each step, both the forward pass and the subsequent backpropagation are confined to a sub-network for \sys.

For the training of \sys, we first start with a warm up stage, training the model with just one processing module, meaning that there is no switch of parameters between forward passes. The training loss consists of a Cross Entropy Loss (CEL) comparing draft model output token distributions with ground truth tokens and a Mean Squared Error (MSE) aligning the target model produced hidden states and the draft model produced hidden states.
$$
    L_1 = \sum_{t=1}^T \textrm{CEL}(\textrm{softmax}(LM_{head}(z^{(t)}_i)),y_i)
$$
$$
    L_2 = \sum_{t=1}^T \textrm{MSE}(z^{(t)}_i, z_{i})
$$
$$
    L_{train}=\lambda_1 L_1+ \lambda_2L_2
$$
Where $z_i$ is the $i$-th target model generated hidden states and  $z_i^{(t)}$ is the $i$-th hidden states outputted by the draft model after its $t$-th forward pass for the sample. $y_i$ is the $i$-th ground truth token . $\lambda_1$ and $\lambda_2$ are coefficients.

After the warm up we replicate the trained transformer weights, producing $M$ copies of the trained transformer and then perform a second phase training. At this time we remove the MSE loss and switch transformers between forward passes.

\subsection{Integrating with SGLang}
Although implementing the speculative algorithm in PyTorch~\cite{paszke2019pytorchimperativestylehighperformance} is convenient, the efficacy of speculative decoding could be overestimated because of missing optimizations from the deployment scenario, such as CUDA graphs and continues batching. Therefore, to concretely evaluate the effectiveness, integrating \sys with an efficient inference framework like SGLang is necessary. We implement \sys inside the SGLang engine and orchestrate \sys with CUDA graph, continues batching, and other inference optimizations.

\section{Experiments}
We experiemented \sys with a variety of settings.

\textbf{Target LLMs.} We conduct extensive experiments on two widely-used open-source LLMs, i.e., LLaMA-2-chat-7B and LLaMA-3-Instruct-8B (Abbreviated as LLaMA-2-7B and LLaMA-3-8B below). We choose the models because they serve as target LLMs in most previous works. Though sharing the same hidden state size, LLaMA-2-7B and LLaMA-3-8B differ greatly in many aspects. Firstly, LLaMA-3 builds its vocabulary with Byte Pair Encoding (BPE) while LLaMA-2 adopts the SentencePiece, resulting in very different vocabularies; LLaMA-3 employs the Grouped Query Attention (GQA) while LLaMA-2 does not; They are also trained with very different datasets and training pipelines. Draft model is sensitive to all of these factors.

\textbf{Datasets.} To verify scaling laws and evaluate the scaling ability of various draft model architectures, we collect and preprocess data from several widely-used open-source datasets, including ShareGPT, UltraChat~\cite{ding2023enhancingchatlanguagemodels}, and OpenThoughts2~\cite{guha2025openthoughtsdatarecipesreasoning}. We preprocess the data by truncating each sample to a maximum length of 2048 tokens and removing content between think tags for samples from OpenThoughts2. Approximately 800,000 samples are drawn from these three datasets. The constitution of the dataset is shown in Table \ref{tab:traindata}. For a fair comparison, the baseline methods mentioned in this paper are also trained on this same dataset.

\begin{table}[tb]
\caption{Constitutions of the training data}
\label{tab:traindata}
\centering
\scriptsize
\vskip 0.15in
\begin{center}

\begin{tabular}{lccc}
\toprule
Dataset & Volume & Proportion & Topic Coverage \\
\midrule
ShareGPT & 68K & 8.5\% & QA/Code/Math/Logic\\
UltraChat & 463K & 57.9\% & QA\\
OpenThoughts2 & 269K & 33.6\% & Math/Logic\\
\bottomrule
\end{tabular}

\end{center}
\vskip -0.1in
\end{table}

\begin{table*}[tb]
\caption{Acceptance lengths (AL) and throughput (TPS, tokens per second) of different methods on NVIDIA A800 GPU}
\vskip 0.15in
\label{tab:exp1}
\centering
\scriptsize
\begin{threeparttable}
\setlength{\tabcolsep}{6pt}
\begin{tabular}{lll cc cc cc cc cc cc cc}
\toprule
\multicolumn{3}{c}{} & \multicolumn{2}{c}{MT-bench} & \multicolumn{2}{c}{HumanEval} & \multicolumn{2}{c}{GSM8K} & \multicolumn{2}{c}{Alpaca} & \multicolumn{2}{c}{CNN/DM} & \multicolumn{2}{c}{Natural Ques.}\\
\cmidrule(lr){4-5} \cmidrule(lr){6-7} \cmidrule(lr){8-9} \cmidrule(lr){10-11} \cmidrule(lr){12-13} \cmidrule(lr){14-15}
Model & Method & Temp & AL & TPS & AL & TPS & AL & TPS & AL & TPS & AL & TPS & AL & TPS\\
\midrule
\multirow{10}{*}{LLaMA-2} & \multirow{2}{*}{Vanilla\tnote{*}} & T = 0 &  N/A  & 95.66  &  N/A  & 96.10  &  N/A  & 96.51  &  N/A  & 96.75  &  N/A  & 92.77  &  N/A  & 96.62  \\
& & T = 1 &  N/A & 95.70  & N/A  & 96.08  & N/A  & 96.31  & N/A  & 96.51 & N/A  & 92.66 & N/A  & 96.447  \\
\cmidrule(l){2-15}
& \multirow{2}{*}{Standard\tnote{**}} & T = 0 & 2.68  & 142.35  & 2.70 & 145.65 & 2.88 & 155.23 & 2.88 & 155.49 & 2.17 & 109.92 & 2.83 & 152.58\\
& & T = 1 & 2.68  & 142.37  & 2.55 & 136.65 & 2.77 & 146.76 & 2.58 & 137.17 & 2.12 & 104.46 & 2.61 & 138.58 \\
\cmidrule(l){2-15}
& \multirow{2}{*}{EAGLE-2} & T = 0 & 4.16 & 246.18 & 4.78 & 283.84 & 4.29 & 254.13 & 4.08 & 243.54 & 3.88 & 218.83 & 3.80 & 225.49 \\
& & T = 1 & 3.87 & 223.69 & 4.49 & 260.25 & 4.26 & 246.24 & 3.87 & 225.22 & 3.70 & 210.09 & 3.60 & 208.45 \\
\cmidrule(l){2-15}
& \multirow{2}{*}{HASS} & T = 0 & 4.48 & 265.08 & 5.15 & 305.31 & 4.54 & 268.94 & 4.42 & 263.70 & 4.18 & 235.41 & 4.08 & 241.61 \\
& & T = 1 & 4.12 & 238.13 & 4.77 & 276.39 & 4.43 & 256.26 & 4.21 & 244.44 & 3.99 & 226.91 & 3.79 & 219.49 \\
\cmidrule(l){2-15}
& \multirow{2}{*}{\textbf{\sys(ours)}} & T = 0 & \textbf{4.66} & \textbf{274.32} & \textbf{5.33} & \textbf{316.16} & \textbf{4.73} & \textbf{279.14} & \textbf{4.67} & \textbf{277.35} & \textbf{4.40} & \textbf{255.08} & \textbf{4.24} & \textbf{250.31} \\
& & T = 1 & \textbf{4.30} & \textbf{246.64} & \textbf{4.90} & \textbf{284.43} & \textbf{4.68} & \textbf{269.19} & \textbf{4.39} & \textbf{254.15} & \textbf{4.15} & \textbf{231.13} & \textbf{4.02} & \textbf{229.80}\\
\midrule
\multirow{10}{*}{LLaMA-3} & \multirow{2}{*}{Vanilla\tnote{*}} & T = 0 &  N/A  & 85.63  & N/A  & 85.88  & N/A  & 85.98  & N/A  & 86.19  & N/A  & 83.18  & N/A  & 86.12  \\
& & T = 1 &  N/A  & 85.39  & N/A  & 85.70  & N/A  & 85.85  & N/A  & 85.79  & N/A  & 82.90  & N/A  & 85.86   \\
\cmidrule(l){2-15}
& \multirow{2}{*}{Standard\tnote{**}} & T = 0 & 5.06 & 164.13 & 5.77 & 180.66 & 5.47 & 173.28 & 4.56 & 144.51 & 4.43 & 143.18 & 4.30 & 136.81 \\
& & T = 1 & 4.42 & 142.69 & 5.45 & 169.80 & 5.20 & 163.19 & 4.16 & 131.12 & 	4.17 & 124.41 & 3.77 & 119.45  \\
\cmidrule(l){2-15}
& \multirow{2}{*}{EAGLE-2} & T = 0 & 3.67 & 168.26 & 4.57 & 212.36 & 4.24 & 197.66 & 3.67 & 171.89 & 3.59 & 163.34 & 3.14 & 147.74 \\
& & T = 1 & 3.67 & 168.64 & 4.30 & 182.15 & 4.00 & 172.00 & 3.37 & 146.32 & 3.30 & 137.69 & 2.99 & 129.65 \\
\cmidrule(l){2-15}
& \multirow{2}{*}{HASS} & T = 0 & 3.93 & 180.50 & 5.12 & 235.38 & 4.73 & 216.75 & 3.94 & 184.39 & 3.90 & 176.16 & 3.36 & 156.67 \\
& & T = 1 & 3.93 & 181.14 & 4.84 & 206.69 & 4.49 & 191.51 & 3.62 & 155.73 & 3.58 & 147.01 & 3.08 & 133.46 \\
\cmidrule(l){2-15}
& \multirow{2}{*}{\textbf{\sys(ours)}} & T = 0 & \textbf{4.29}	& \textbf{201.51} &	\textbf{5.43} & \textbf{251.96}	& \textbf{5.11} & \textbf{237.41} &	\textbf{4.25} &	\textbf{199.31} & \textbf{4.17} &	\textbf{187.95} & \textbf{3.63} & \textbf{170.06} \\
& & T = 1 & \textbf{4.29} &	\textbf{201.47} & \textbf{5.02} &	\textbf{216.30} & \textbf{4.76}	& \textbf{205.09} &	\textbf{3.86} &	\textbf{167.72} &	\textbf{3.76} &	\textbf{158.33} &	\textbf{3.31} &	\textbf{143.63} \\
\bottomrule
\end{tabular}
\begin{tablenotes}
    \item[*] \scriptsize Vanilla means the standard auto-regressive decoding. Thus, there is no acceptance length for vanilla decoding.
    \item[**] \scriptsize Standard means using a light weight version of the same series models as the drafter. For LLaMA-2-7B we employ the LLaMA-160M model while for LLaMA-3-8B we employ the LLaMA-3.1-1B model. Same with the experiments on 4090 GPUs.
\end{tablenotes}
\end{threeparttable} 
\end{table*}
\textbf{Benchmarks.} Following previous works, we evaluate on six benchmarks,i.e.,MT-Bench~\cite{chen2025mtbenchmultimodaltimeseries},HumanEval~\cite{chen2021evaluatinglargelanguagemodels},GSM8K~\cite{cobbe2021trainingverifierssolvemath}, Alpaca~\cite{alpaca}, CNN/Daily Mail~\cite{see-etal-2017-get} and Natural Questions~\cite{kwiatkowski-etal-2019-natural}, each representing very different workloads. MT-bench is for multi-round conversation; HumanEval is for code completion; GSM8k evaluates LLM on grade-school-level math; Alpaca is a dataset for instrution following; CNN/Daily Mail is for summarization tasks and Natural Questions consists of real questions the users search on google.

\textbf{Metrics.} We focus on the following metrics to evaluate draft models:
\begin{itemize}
    \item \textbf{Acceptance Length}. Acceptance Length is the expected number of draft tokens that the target model accepts per verification step. It reflects how good the draft model predicts target model output.
    \item \textbf{Throughput}. Throughput, measured in tokens per second, measures the end-to-end rate at which a speculative decoding system generates tokens. It is affected not only by the acceptance length but also by the computational overhead.
\end{itemize}

\textbf{Baselines.} We compare \sys with various baseline drafter models, including the standard speculative decoding, EAGLE-2, HASS and EAGLE-3.

\textbf{Train Setting.} We train \sys and all baseline drafters on datasets rangeing from 100,000 to 800,000 samples with 8 NVIDIA A100 40G GPUs and 4 NVIDIA A100 80G GPUs. Generally, more epochs are trained for smaller datasets and less epochs are trained for larger datasets. The time cost for the trainings vary on dataset sizes, ranging from 1 day to 2 weeks.  

Note that, \sys evaluated in the main experiments consists of two processing modules. While the first one is assigned uniquely to process the first prefill step, the second is responsible for all subsequent decoding steps. 

\subsection{Effectiveness}
\begin{table*}[tb]
\caption{Acceptance lengths (AL) and throughput (TPS, tokens per second) of different methods on NVIDIA 4090 GPUs}
\vskip 0.15in
\label{tab:exp2}
\centering
\scriptsize
\setlength{\tabcolsep}{6pt}
\begin{tabular}{lll cc cc cc cc cc cc cc}
\toprule
\multicolumn{3}{c}{} & \multicolumn{2}{c}{MT-bench} & \multicolumn{2}{c}{HumanEval} & \multicolumn{2}{c}{GSM8K} & \multicolumn{2}{c}{Alpaca} & \multicolumn{2}{c}{CNN/DM} & \multicolumn{2}{c}{Natural Ques.}\\
\cmidrule(lr){4-5} \cmidrule(lr){6-7} \cmidrule(lr){8-9} \cmidrule(lr){10-11} \cmidrule(lr){12-13} \cmidrule(lr){14-15}
Model & Method & Temp & AL & TPS & AL & TPS & AL & TPS & AL & TPS & AL & TPS & AL & TPS\\
\midrule
\multirow{10}{*}{LLaMA-2} & \multirow{2}{*}{Vanilla} & T = 0 & N/A & 101.30  & N/A & 103.57  & N/A & 104.25 & N/A & 103.89 & N/A & 99.73 & N/A & 	104.52    \\
& & T = 1 & N/A & 102.27 & N/A & 103.27 & N/A & 102.54 & N/A & 103.80  &	N/A & 99.60 & N/A  & 103.38     \\
\cmidrule(l){2-15}
& \multirow{2}{*}{Standard} & T = 0 & 2.77 & 134.85 & 2.72 & 133.32 & 2.88 & 140.77 & 2.90 & 143.16 & 2.05 & 96.59 & 2.84 & 136.31\\
& & T = 1 & 2.78 & 136.56 & 2.62 & 129.76 & 2.85 & 138.21 & 2.85 & 137.00 & 2.02 & 94.20 & 2.79 & 134.66 \\
\cmidrule(l){2-15}
& \multirow{2}{*}{EAGLE-2} & T = 0 & 4.12 & 223.32 & 4.75 & 251.94 & 4.26 & 222.95 & 4.09 & 217.64 & 3.89 & 197.64 & 3.84 & 201.01 \\
& & T = 1 &4.12   &223.28 & 4.48 & 218.69 & 4.24 & 205.44 & 3.92 & 192.05 & 3.71 & 174.54 & 3.64 & 173.72  \\
\cmidrule(l){2-15}
& \multirow{2}{*}{HASS} & T = 0 & 4.42 & 	238.65& 5.12 & 271.59 & 4.51 & 238.27 & 4.40 & 233.42 & 4.20 & 213.89 & 4.08 & 217.04 \\
& & T = 1 &4.43 & 241.22  & 4.74 & 233.54 & 4.50 & 218.77 & 4.18 & 207.23 & 4.01 & 190.18 & 3.78 & 179.97 \\
\cmidrule(l){2-15}
& \multirow{2}{*}{\textbf{\sys(ours)}} & T = 0 & \textbf{4.69} & \textbf{254.18} & \textbf{5.26} & \textbf{287.77} & \textbf{4.68} & \textbf{256.28} & \textbf{4.62} & \textbf{256.10} & \textbf{4.35} & \textbf{217.02} & \textbf{4.24} & \textbf{232.58}  \\
& & T = 1 & \textbf{4.65} & \textbf{253.61} & \textbf{4.88} & \textbf{250.55} & \textbf{4.63} & \textbf{234.73} & \textbf{4.34} & \textbf{223.28} & \textbf{4.18} & \textbf{194.20} & \textbf{3.96} & \textbf{195.25} \\
\midrule
\multirow{10}{*}{LLaMA-3} & \multirow{2}{*}{Vanilla} & T = 0 & N/A  & 91.60  & N/A  & 91.51  & N/A  & 91.42  & N/A  & 91.32  & N/A  & 88.51  & N/A  & 90.88   \\
& & T = 1 & N/A & 91.72 & N/A & 91.02 & N/A & 90.83 & N/A & 91.01 & N/A & 88.46 & N/A & 90.97    \\
\cmidrule(l){2-15}
& \multirow{2}{*}{Standard} & T = 0 & 5.08 & 149.47  & 5.93 & 171.76 & 5.75  & 166.41 & 4.80 & 139.88 & 4.75 & 135.04 & 4.48  & 130.60 \\
& & T = 1 & 5.08 & 150.61 & 5.36 & 154.60 & 5.31 & 152.54 & 4.28 & 124.80 & 4.30 & 121.33 & 3.93 & 115.29 \\
\cmidrule(l){2-15}
& \multirow{2}{*}{EAGLE-2} & T = 0 & 3.76 & 166.75 & 4.57 & 198.19 & 4.26 & 	186.82 & 3.63 & 160.91 & 3.58 & 153.15 & 3.24 & 142.68  \\
& & T = 1 & 3.77 & 168.10 & 4.35 & 154.91 & 4.02 & 142.84 & 3.37 & 120.82 & 3.34 & 115.65 & 0.98 & 107.00  \\
\cmidrule(l){2-15}
& \multirow{2}{*}{HASS} & T = 0 & 4.01 & 184.12 & 5.16 & 230.42  & 4.74 & 212.79 & 3.91  & 178.80 & 3.93 & 170.92 & 3.41 & 155.38  \\
& & T = 1 & 4.01 & 183.37 & 4.75 & 169.27 & 4.45  & 157.93 & 3.58 & 127.26 & 3.58  & 124.07 & 3.13 & 112.40 \\
\cmidrule(l){2-15}
& \multirow{2}{*}{\textbf{\sys(ours)}} & T = 0 & \textbf{4.27} & \textbf{198.21} & \textbf{5.44} & \textbf{252.18} & \textbf{5.06} & \textbf{235.85} & \textbf{4.21} & \textbf{198.16} & \textbf{4.15} & \textbf{178.01} & \textbf{3.58} & \textbf{168.44}  \\
& & T = 1 & \textbf{4.30} & \textbf{198.65}  & \textbf{5.08} & \textbf{182.53} & \textbf{4.71} & \textbf{170.26} & \textbf{3.85} & \textbf{139.94} & \textbf{3.71} & \textbf{125.98} & \textbf{3.21} & \textbf{117.48}  \\
\bottomrule
\end{tabular}
\end{table*}

Table \ref{tab:exp1} shows our main results, comparing \sys with baseline methods on six workloads, two target models and under greedy and non-greedy sampling settings. The experiments simulate the condition where 1 NVIDIA A800 80G GPU is used for target LLM generation with speculative sampling. An extra set of experiments on 2 NVIDIA 4090 24G GPUs can be found in Table~\ref{tab:exp2}. All experiments are conducted with batch size 1. A 6-step, 4-branch tree structure is used for the tree-shape draft and verification. A maximum of 16 tokens are validated for each verification. Consistent advantages over all baseline methods in all testing settings is observed. \sys consistently gains more than 2.4x acceleration compared to the standard vanilla auto-regressive decoding . Compared to other baseline drafter models that activate the same amount of parameters during inference, \sys on average improves acceptance length by 14.09\% and 5.69\% and generate 14.21\% and 6.10\% more tokens per second than EAGLE-2 and HASS, respectively. Speculative decoding performance is task dependent. \sys works exceptionally well, compared to other baselines on CNN/Daily Mail work load, outperforms HASS by potentially suggests the step splitting architecture may work better for long context tasks. 

\subsection{Re-examine Scaling Laws of \sys}
We conduct experiments to verify the drafter scaling phenomenon observed in previous works~\cite{li2025eagle3scalinginferenceacceleration, yan2025scalinglawsspeculativedecoding}. We also prove with these experiments that even without increasing the number of activated parameters per prediction, the model's capacity expands as the volume of training data grows. This expansion is evidenced by a clear improvement in predictive performance. To more clearly demonstrate the upper bound of predictive power for various draft models at different training data scales, we employed a more extreme tree structure in the experiments: a 6-step, 10-branch tree, with each step verifying up to 60 tokens.

\begin{figure}[t]
    \centering
    \includegraphics[width=\linewidth]{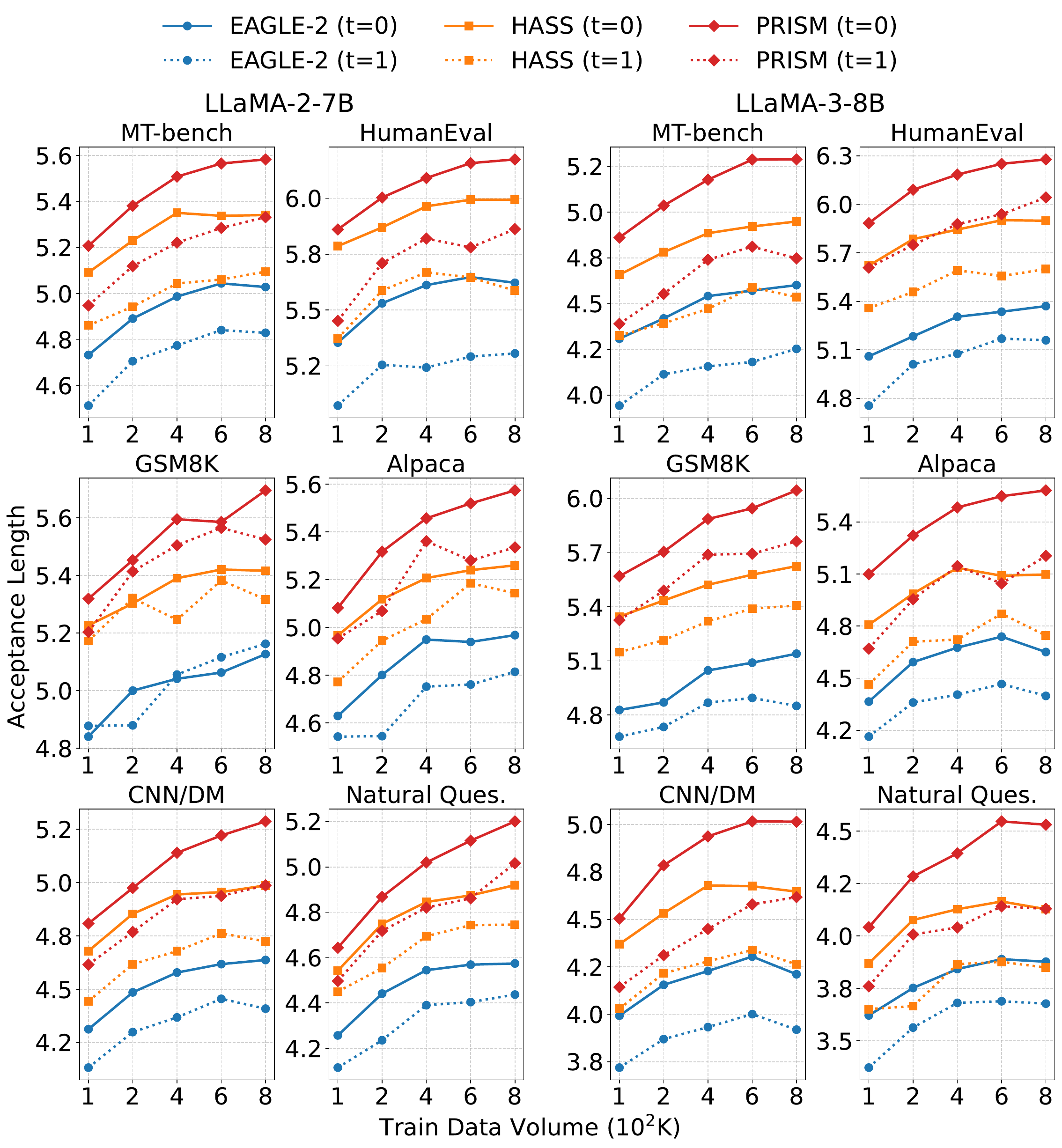}
    \caption{Scaling curves of \sys and comparative drafters over 6 benchmarks and greedy and non-greedy sampling.}
    \label{fig:scaling}
\end{figure}

The first results are presented in Figure~\ref{fig:scaling}, comparing \sys with drafters that activate a similar amount of parameters during inference, i.e., EAGLE-2 and HASS. Across all tested data scales, \sys demonstrates a big advantage in acceptance length over both EAGLE-2 and HASS. 

We further analyze the curves. First, in the low-data regime with fewer than 200,000 training samples, the representational capacity of the single-transformer models has not yet been saturated. This is indicated by their performance continuing to improve with more data. Nevertheless, \sys already outperforms both baselines at this scale. This supports our hypothesis that the hierarchical structure of \sys progressively generates more informative representations as data passes through its stacked processing modules, leading to better predictions. Second, in the high-data regime, corresponding to the latter half of the curves, a clear scaling bottleneck becomes evident for both EAGLE-2 and HASS. While \sys still shows potential to scale with more than 600,000 samples, the performance of EAGLE-2 and HASS begins to plateau after 400,000 samples, respectively. This demonstrates the superior scaling ability of \sys compared to single-transformer draft models like EAGLE-2 and HASS.

We also compare \sys against EAGLE-3, an architecture specifically designed for scaling. EAGLE-3 expands attention layer parameters and leverages multiple target LLM produced hidden states, all of which \sys does not equip. 

\begin{figure}[tb]
\centering
\begin{minipage}[t]{0.5\linewidth}
\centering
\includegraphics[width=\linewidth]{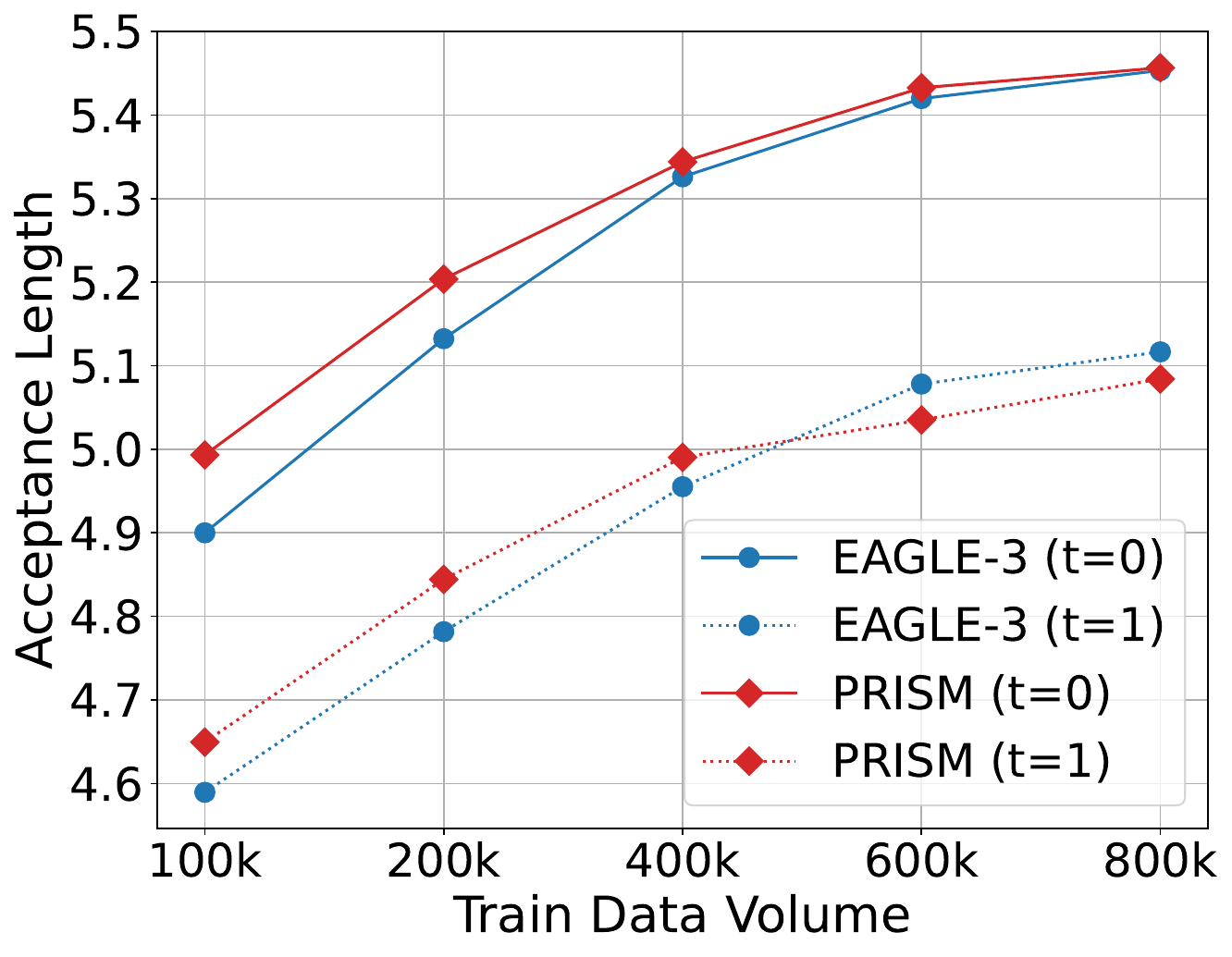}
\caption{\sys vs. EAGLE-3}
\label{fig:eagle3cmp}
\end{minipage}\hfill
\begin{minipage}[t]{0.5\linewidth}
\centering
\includegraphics[width=\linewidth]{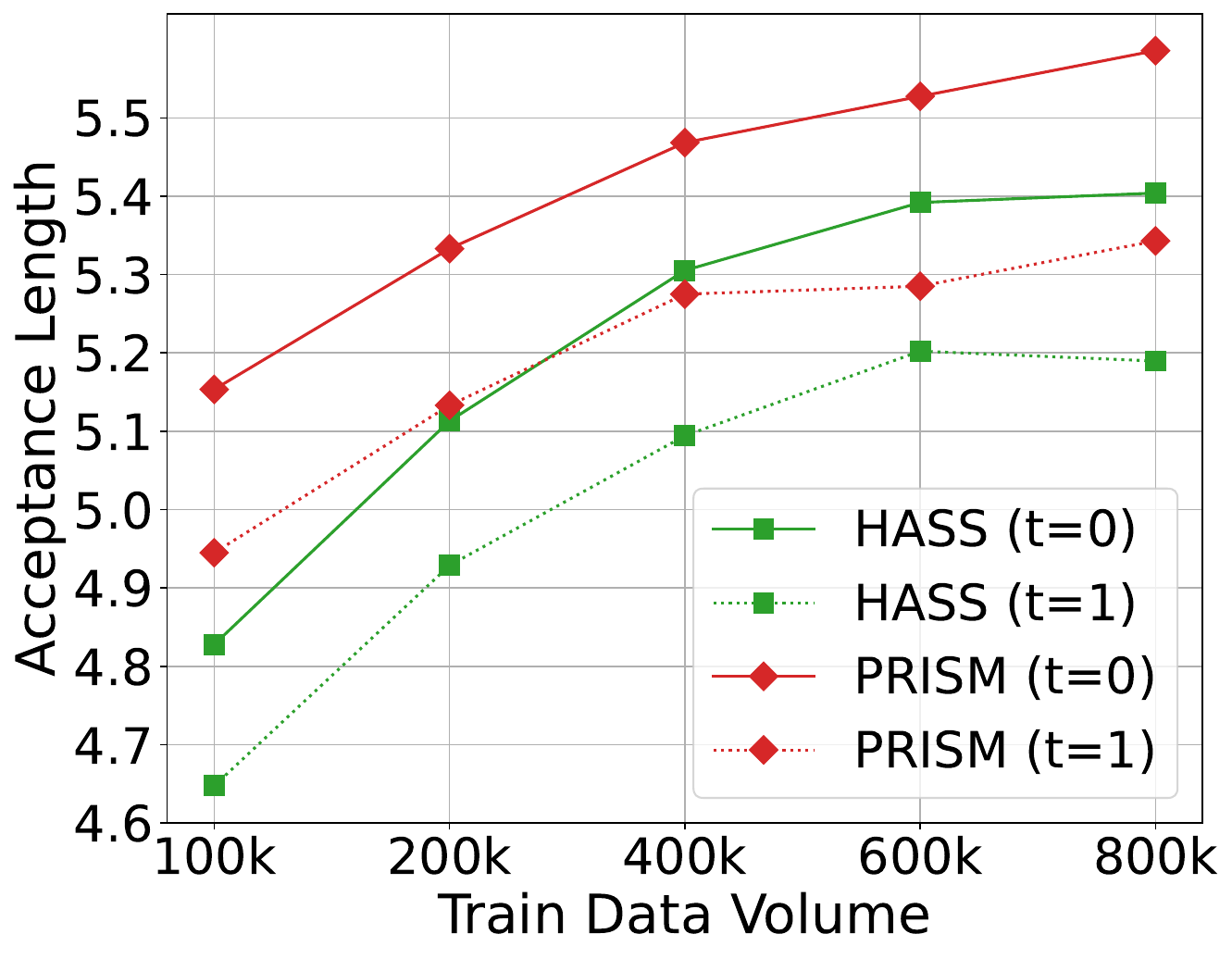}
\caption{Ablation for \sys}
\label{fig:ablation}
\end{minipage}
\end{figure}

Figure~\ref{fig:eagle3cmp} compares the average scaling ablity of \sys and EAGLE-3, when trained as drafters for LLaMA-3-8B. \sys demonstrates better data efficiency, outperforming EAGLE-3 when trained with relatively small amounts of data. As the training data volume increases, both methods converge to the same acceptance length. We assume this is due to the limited complexity of our dataset. Although they feature a similar predictive performance, \sys and EAGLE-3 behave very differently on a step-by-step basis. As shown in Figure~\ref{fig:ar}, \sys predicts better for initial steps, while EAGLE-3 predicts better for later steps. Both methods have a gentler slope compared to EAGLE-2 and HASS. For EAGLE-3, this can be attributed to its train-time test context alignment and more informative inputs, while for \sys, it is attributed to our step-wise specialized parameters.

An ablation experiment is also performed to prove that the scaling efficacy directly comes from the core innovation of \sys: distributing draft step computation to different sets of parameters. We conduct this experiment by compare \sys with a model directly stacking two transformers. We train the comparative model with exactly the same context alignment and dataset with our \sys instance. The result is shown in Figure \ref{fig:ablation}. Without our core innovation, though powered with more parameters, the ablation model predicts and scales considerably bad than \sys.

\subsection{Hyper-Parameters}

Speculative decoding is a complex system where lots of hyper parameters influences the end-to-end acceleration. In this section, we demonstrate these influences, revealing insights for achieving higher end-to-end throughput.

\begin{figure}[tb]
\centering
\begin{minipage}[t]{0.47\linewidth}
\centering
\includegraphics[width=\linewidth]{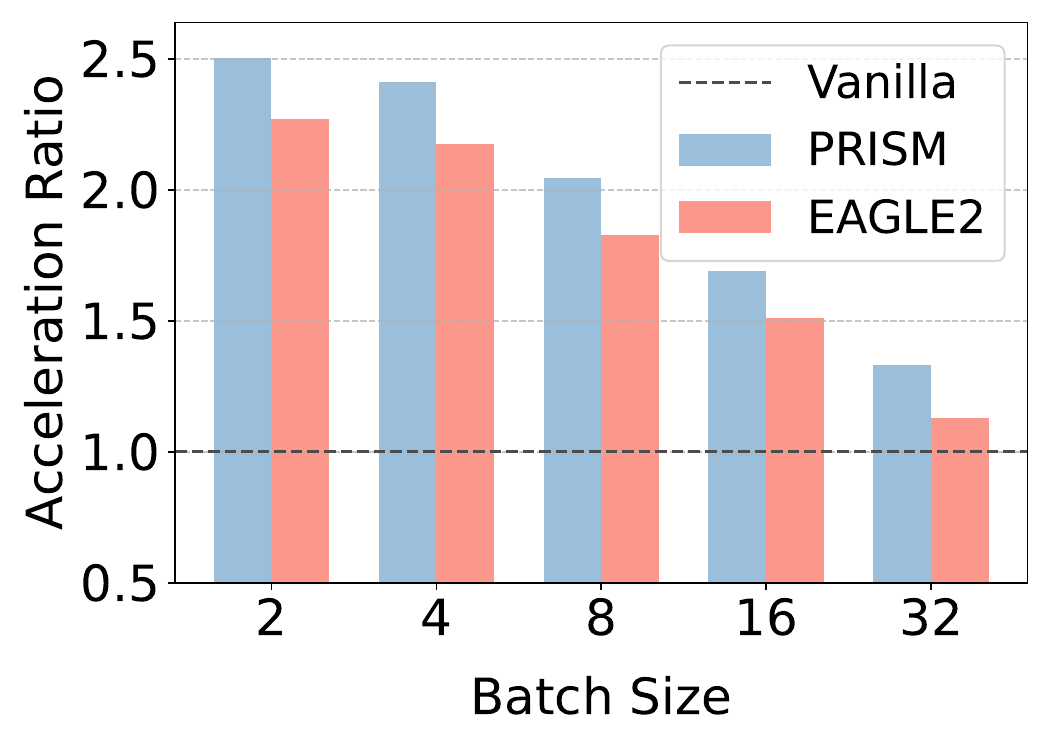}
\caption{A comparision between \sys and EAGLE-2 acceleration with larger batch size}
\label{fig:large batch}
\end{minipage}\hfill
\begin{minipage}[t]{0.5\linewidth}
\centering
\includegraphics[width=\linewidth]{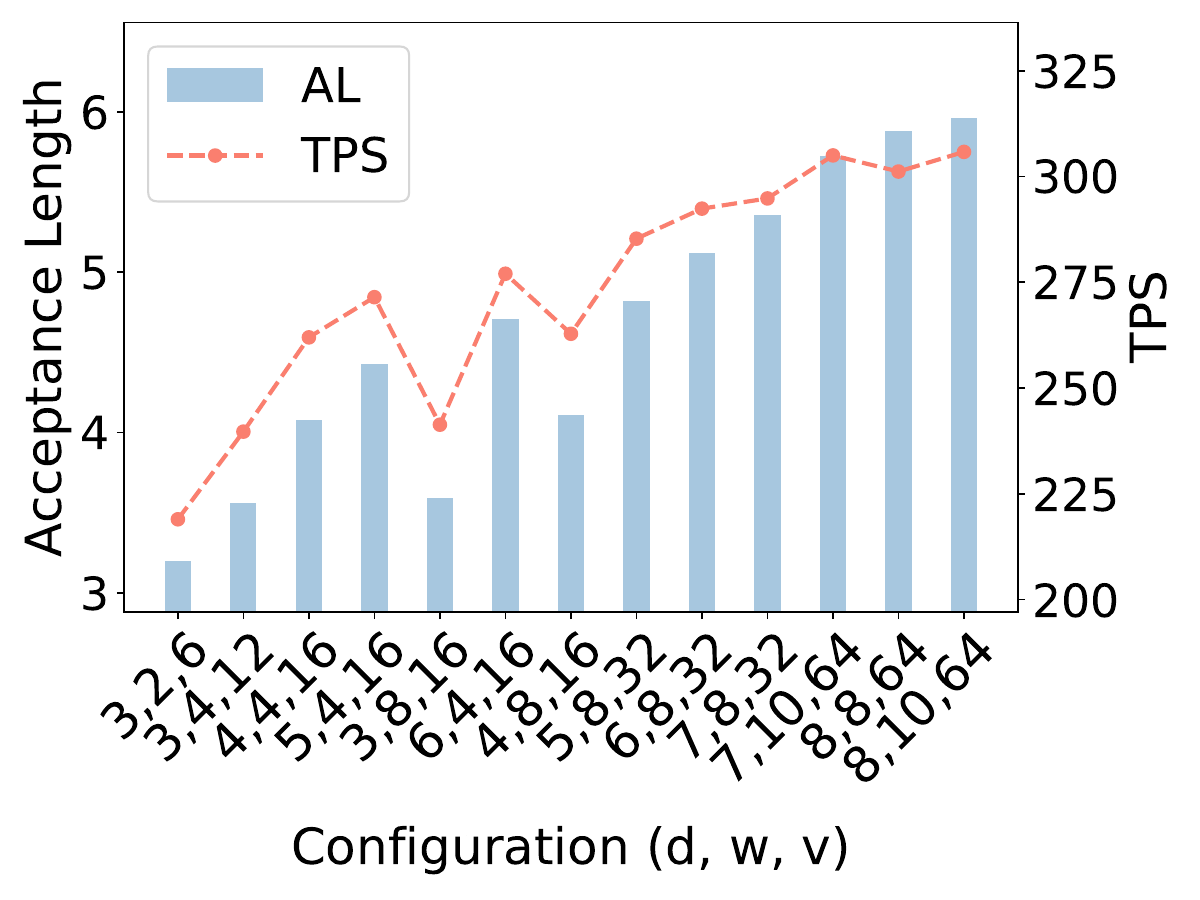}
\caption{\sys acceptance length and throughput under different tree topologies.}
\label{fig:tree}
\end{minipage}
\end{figure}

\textbf{Batch Size}.
The benefit of speculative sampling diminishes with increasing batch sizes, as larger batch sizes make LLM decoding more compute-intensive and less constrained by memory bandwidth. It is often a trade-off between employing speculative sampling and utilizing big batch sizes. Nonetheless, \sys still provides reasonable speedup when evaluated with relatively larger batch sizes, as shown in Figure~\ref{fig:large batch}.

\textbf{Tree Topology}.
In speculative sampling, the tree topology defines the depth (d) and width (w) of the tree of candidate tokens generated by the draft model. To keep the computational cost of the validation step manageable, the total number of verified nodes (v) is typically fixed. The optimal tree topology, however, is highly dependent on the predictive characteristics of the draft model. For instance, a draft model capable of accurate long-range predictions may perform best with a deeper, narrower tree. Conversely, a model whose predictive accuracy drops off more quickly may benefit from a wider, shallower tree that explores more parallel branches. In Figure~\ref{fig:tree}, we explore the effect tree topology has on \sys.

Experimental results indicate that employing a larger tree generally enhances the performance of \sys. However, speedups improve with growing tree depth and width gradually saturate. This is due to the growing overhead of both draft and verify stages coming with larger trees. Specifically, for a fixed number of tokens to be verified, a deeper, narrower tree structure outperforms a shallower, wider one.

\section{Related Works}

Apply speculative sampling techniques to accelerate LLMs was first introduced by \citealt{leviathan2023fast}. Since that, extensive efforts have been made from both the system and AI research community to improve the framework.

Efforts have been made exploring speculative sampling in LLM serving systems. SpecInfer \cite{10.1145/3620666.3651335} firstly introduce the tree-based speculative inference and token verification mechanism instead of an incremental decoder, which significantly reduces the verification overhead.  SwiftSpec \cite{zhang2025swiftspecultralowlatencyllm} finds the imbalanced compute requirements for draft and target model under the common circumstance that the draft and target model use the same tensor parallel configuration. It redesigns the speculative decoding pipeline in an asynchronous and disaggregated method, thus assign the computation resources by requirements. AdaServe \cite{li2025adaserveacceleratingmultislollm} supports multi-SLO requirements through SLO-customized speculative decoding. It improves the goodput and reduces SLO violations by the hardware-aware algorithm and speculate-select-verify pipeline. Besides, because reinforcement learning including rollout becomes a pivotal methodology for enhancing LLMs and imbalances in rollout lengths cause low GPU utilization, speculative sampling is also a promising method to alleviate this problem and accelerate RL training \cite{he2025historyrhymesacceleratingllm}.

Another significant body of work has focused on improving the efficiency of speculative decoding by increasing the number of accepted tokens per step. Medusa \cite{pmlr-v235-cai24b} pioneered the multi-head approach for draft generation. Instead of using a separate draft model, Medusa augments a pretrained backbone by attaching multiple, lightweight decoding heads to its final layer. The EAGLE series \cite{li2024eagle,li2024eagle2fasterinferencelanguage,li2025eagle3scalinginferenceacceleration} further improves acceptace length by feeding richer feature representations from the target model into the draft heads. This enhanced context leads to higher acceptance rates and, consequently, greater end-to-end speedups. EAGLE-2 introduced a dynamic tree-building algorithm, while EAGLE-3 further improved performance by incorporating a context alignment technique during training and leveraging additional hidden states from the target model. HASS \cite{zhang2025learning} independently developed its own version of context alignment, proposing a novel two-phase training scheme and a distinct loss function to optimize. Focusing on the same challenge of context mismatch, Griffin \cite{hu2025griffineffectivetokenalignment} proposed that not only hidden states but also the input tokens themselves should be synchronized between the draft and target models during training and inference. Shifting the focus, Jakiro \cite{huang2025jakiroboostingspeculativedecoding}, notable for its performance in non-greedy sampling scenarios, introduced MOE and contrastive learning into the draft model training. 

Finally, the challenge of scaling draft models was explored by Scylla \cite{yan2025scalinglawsspeculativedecoding}, which investigated the impact of increasing the parameter count of the drafters. A concurrent work from Meta \cite{tang2025efficientspeculativedecodingllama} independently conducted similar experiments, confirming that larger, more capable drafters can yield substantial performance gains.

\section{conclusion}

In this paper, we introduce \sys, a novel architecture for draft model of speculative sampling. \sys deals with the central trade-off between draft model effectiveness and efficiency. By parametrically disaggregating the computation of draft steps, draft model capacity increases while computation cost for each step maintains. Extensive experiments are performed, verifying scaling law for draft models while proving effective scaling for the \sys architecture. We also demonstrate the practicability of \sys in real-life scenarios by integrating it into sglang and proving the effectiveness.


\bibliography{example_paper}
\bibliographystyle{mlsys2025}

\appendix

\end{document}